# Investigating Post-pretraining Representation Alignment for Cross-Lingual Question Answering

**Fahim Faisal, Antonios Anastasopoulos**
Department of Computer Science, George Mason University
{ffaisal,antonis}@gmu.edu

## Abstract

Human knowledge is collectively encoded in the roughly 6500 languages spoken around the world, but it is not distributed equally across languages. Hence, for information-seeking question answering (QA) systems to adequately serve speakers of all languages, they need to operate cross-lingually. In this work we investigate the capabilities of multilingually pretrained language models on cross-lingual QA. We find that explicitly aligning the representations across languages with a post-hoc fine-tuning step generally leads to improved performance. We additionally investigate the effect of data size as well as the language choice in this fine-tuning step, also releasing a dataset for evaluating cross-lingual QA systems.[1]

## 1 Introduction

Information seeking question answering, where a user asks a question to get a related passage or short text as answer, is a widely studied area (Clark et al., 2020; Kwiatkowski et al., 2019; Yang et al., 2015, *inter alia*)[2] that has been successfully deployed in user-facing applications such as conversational assistants (Gao et al., 2018). For example, an English computer science student that asks Apple Siri, Amazon Alexa, or Google Assistant the question "Where did Joan Clarke work?"[3] will receive the answer "Bletchley Park", an answer based on the English Wikipedia entry for Joan Clarke automatically retrieved by the system.

Now, consider another student, this time based in Greece or Bangladesh, asking effectively the same question, "Πού δούλευε η Joane Clarke;" or "Joane Clarke কোথায় কাজ করত?", but this time in their native language Greek or Bengali. However, as of July 2021, there is no Wikipedia article for Joane Clarke in Greek or in Bengali.[4] For the QA system to adequately serve these students, it will need to function in a cross-lingual setting, retrieving the English (or any of the other available languages) article and producing an answer given the question in a different language (Asai et al., 2021). Throughout this paper we will refer to a setting where the question and the context are in different languages as "cross-lingual" QA.

Multilingually-pretrained language models such as mBERT (Devlin et al., 2019) or XLM-R (Conneau et al., 2020) are widely used as the base of modern QA systems and have shown promise for performing the task in zero-shot (Artetxe et al., 2020) or few-shot manner (Debnath et al., 2021) as well as in cross-lingual settings (Asai et al., 2021). At the same time, these models are not without drawbacks; their pre-training objectives did not explicitly require that their representations *align* across languages for semantically similar words/phrases/sentences, with adverse effects especially for languages written in scripts other than the Latin one (Muller et al., 2020). A recent line of work (Cao et al., 2020; Zhao et al., 2021; Kulshreshtha et al., 2020; Huang et al., 2021) in fact shows improvements on a variety of natural language tasks by explicitly or implicitly aligning the models' representations across languages leveraging parallel data.

In this work, we explore the effect of such representation alignment techniques on the performance of QA systems in cross-lingual settings, concluding that they can be very beneficial. Our contributions can be summarized as follows:

1. We augment the TyDi-QA dataset (Clark et al., 2020) to study cross-lingual QA settings, using

[1] Code and dataset are publicly available here: https://github.com/ffaisal93/aligned_qa

[2] See (Rogers et al., 2021) for a thorough survey of the field.

[3] Joan Clarke was the only female computer scientist at Bletchley Park during the Second World War. Details: https://en.wikipedia.org/wiki/Joan_Clarke

[4] See (Jiang et al., 2020) for a visualization of the different Wikipedia sizes across languages (Figure 1) and the distribution of informative facts across languages (Figure 5).

both synthetic and newly-collected human translations.
2. We show that alignment-based fine-tuning of pretrained language models, along with data augmentation of the training data leads to improvements of 21.1% on average across TyDi-QA primary tasks in cross-lingual settings.
3. We perform thorough analyses to find strategies for selecting languages and training data for this multilingual alignment fine-tuning step.

## 2 Cross-lingual TyDi-QA dataset

To effectively study the cross-lingual capabilities of extractive QA systems, we construct a dataset with questions and potential contexts in different languages. We base our cross-lingual QA dataset on the TyDi-QA dataset (Clark et al., 2020). TyDi-QA dataset covers 11 languages with 204k question-answer pairs. Unlike other multilingual datasets like MLQA (Lewis et al., 2020) and XQuAD (Artetxe et al., 2020), TyDi-QA dataset is collected from native speakers without the aid of translation.

We select 5 typologically diverse language: Bengali, Korean, Arabic, Swahili, and English from TyDi-QA dataset for all of our experiments. Similar to other TyDi-QA-derivative datasets we use the publicly available development set as our test set, and sample a few examples from the training data to build a custom development set (these are examples are discarded from the training set). We will refer to this (original) dataset as TyDi-QA-`mono` to denote that all training/test instances are in a monolingual setting. We augment the dataset both automatically by generating translations of questions and by collecting human-created translations for some test sets. We describe these two datasets below:

**Synthetic Translations** We create a cross-lingually augmented version of our training set in two ways. First, we augment the English portion by automatically translating English questions to the other four languages ($Q_{en} \xrightarrow{\texttt{translate}} Q_{ar,sw,ko,bn}$), keeping the context (and answers) in English. Similarly, we augment the Bengali, Swahili, Korean, and Arabic portions of the dataset by translating their questions to english ($Q_{ar,sw,ko,bn} \xrightarrow{\texttt{translate}} Q_{en}$), still keeping the context and answers in the original language.[5] We will refer to this version of the dataset as TyDi-QA+`aug`. For all translations we use the Google Translate API[6] but we will make them publicly available to ensure the reproducibility of our experiments. The same process is applied to our custom development and test sets.

**Human Translations** In addition to machine generated translations, we asked native Bengali and Swahili speakers[7] who are also fluent in English to record English translations of the Bengali and Swahili test set questions. The annotators were shown the original Bengali/Swahili questions, then asked to translate these questions to English and recorded. The process is similar to the one used by Faisal et al. (2021) in creating the SD-QA dataset. We then transcribed these recordings using the Google ASR (automatic speech recognition) API.[8]

This allows us to evaluate the models in a more realistic scenario (similar to SD-QA, but now in a cross-lingual setting): a speaker provides an *oral* query to the model, which has to first transcribe the question to be ran through the QA system. This dataset was collected as part of the original SD-QA (Faisal et al., 2021) data collection process. We refer SD-QA dataset (Faisal et al., 2021) for detailed annotation process and data statistics.

## 3 Alignment-based fine-tuning

Language models like mBERT and XLM-R are trained without the inclusion of any parallel data. Cao et al. (2020) use a small amount of parallel data to align the representations of similar words across different languages, an approach that leads to improvements in cross-lingual inference tasks.

This method relies on the availability of $k$ parallel corpora: $c_1, c_2, ...c_k$ where each corpus $c_{(l_s,l_t)} = (\mathbf{s}_1, \mathbf{t}_1), ...(\mathbf{s}_n, \mathbf{t}_n)$ is a set of parallel sentences in the source-target $(l_s, l_t)$ languages. Using any word alignment technique, we can obtain word-level alignments $\alpha(\mathbf{s}, \mathbf{t}) = (p_1, q_1)....(p_m, q_m)$ for each sentence pair, such that for each $(p, q)$ pair, word $s_p$ of source sentence **s** is translated to word $t_q$ in target sentence **t**.

We can devise an objective function that minimizes the distance of the contextualized embeddings of aligned words, by penalizing the distance of their representations. Denoting with $f_\phi(s_i)$ the contextual embedding of word in position $i$ for sentence $\mathbf{s}$ given by a model parameterized by $\phi$, we get

$$L(c_{(l_s,l_t)}, f_\phi) = \sum_{(\mathbf{s},\mathbf{t})\in c_{(l_s,l_t)}} \sum_{(p,q)\in\alpha(\mathbf{s},\mathbf{t})} \|f_\phi(s_p) - f_\phi(t_q)\|_2^2$$

[5] While English is not always the best pivot language for cross-lingual transfer (Anastasopoulos and Neubig, 2020, *inter alia*), in this case it is the speakers' most likely second language.

[6] `https://cloud.google.com/translate`

[7] The annotators were native speakers of the two languages residing in Bangladesh and Kenya.

[8] `https://cloud.google.com/speech-to-text`

This fine-tuning process runs the risk of leading the encoder's parameters too far from their initial values $\phi_0$, which could lead to catastrophic forgetting (French, 1999; Ratcliff, 1990; McCloskey and Cohen, 1989). Hence, following Cao et al. (2020) we additionally use a regularization metric to penalize the excessive drifting from the initial encoder state $\phi_0$ for each of the languages.

$$R(c_{(l_s,l_t)}, f_\phi) = \sum_{\mathbf{s},\mathbf{t}\in c_{(l_s,l_t)}} \Big(\sum_{i=1}^{|\mathbf{s}|} \|f_\phi(s_i) - f_{\phi_0}(s_i)\|_2^2 + \sum_{i=1}^{|\mathbf{t}|} \|f_\phi(t_i) - f_{\phi_0}(t_i)\|_2^2\Big).$$

The final objective is simply the sum of the two components, similar to the approaches for retrofitting static embeddings (Faruqui et al., 2015):

$$\mathbf{min}_\phi \sum_{v=1}^{k} L(c_v, f_\phi) + R(c_v, f_\phi)$$

## 4 Experimental Setup

**Model and Baselines** Our extractive QA model architecture is similar to Alberti et al. (2019). In this implementation, both the question and the context are encoded to contextual representations using pre-trained language models such as BERT, and a dedicated classification head produces answer depending on the two types of tasks: 1) the index of the passage answering the questions or Null (if no such answer exist) for Passage selection task. 2) a Yes/No answer or the answer span using these representations or Null for Minimal answer span task.

We train baseline models using both TyDi-QA-`mono` and TyDi-QA+`aug` training datasets. All models are trained multilingually, training using all languages jointly. We report results on the two TyDi-QA primary tasks, passage selection (given a question and a number of candidate passages, this task is to return the index of the passage containing the answer) and minimal answer selection (a question and a number of passages are given; the task is to return the start and end byte of a short text span containing the answer).

We preform experiments with both mBERT and XLM-R. Our baselines use the pre-trained models, fine-tuned as above on the QA task (without any intermediate alignment-based fine-tuning). Our models, in turn, first perform alignment-based fine-tuning on the pre-trained models, and then train them on the QA task.

**Alignment-Based fine-tuning** Due to the lack of $n$-way parallel corpora, all parallel corpora we use have English as one of the two languages. We obtain word-level alignments using AWESOME-ALIGN (Dou and Neubig, 2021). We use data from WikiMatrix (Schwenk et al., 2021) , Wikimedia (Tiedemann, 2012) and CC-aligned (El-Kishky et al., 2020) to prepare various versions of our parallel corpus. The size of these data varies from 260k to less than 1k for each language based on experiment type. To study the effect of using different language samples and data sizes, we experiment with different models (summarized in Table 1):

- CAO-HIGH: this is the mBERT model provided by Cao et al. (2020), originally fine-tuned on English-X parallel data for five high resource languages: Bulgarian, German, Greek, Spanish, and French, using between 10 to 260 thousand parallel sentences in each pair.
- ZHAO-LOW: this mBERT trained model is from the experiments done by (Zhao et al., 2021). The authors trained different variations of aligned models based on the training language similarity level with English (measured cosine-distance using mBERT). The ZHAO-LOW is trained on 9 language-English parallel corpus having low to medium similarity level with English. These languages are from originally three main language families: 1. Austronesian, 2. Germanic , 3. Indo-Aryan
- ZHAO-(LOW+HIGH): Like ZHAO-LOW, this model is also provided by (Zhao et al., 2021) and is trained on XLM-R. The training languages used here are from all five language families: 1. Austronesian, 2. Germanic , 3. Indo-Aryan 4. Romance, and 5. Uralic to maintain a balance in terms of the similarity level with English.

Beyond these models, we also explore the following settings:

- TYDI-L: we use 80 to 260k parallel sentences only between English and the languages in our

| Model name | Description | Languages | Parallel Data (per lang.) | Base Model mBERT | XLM-R |
|---|---|---|---|---|---|
| CAO-HIGH | from (Cao et al., 2020) | eng–{bul,deu,ell,esp,fra} | 250k | ✓ | |
| ZHAO-LOW | from (Zhao et al., 2021) | eng–{hin,ind,jav,tgl,mar,urd, afr,msa,ben} | 100k | ✓ | |
| ZHAO-(LOW+HIGH) | from (Zhao et al., 2021) | eng–{deu,por,nld,ind,ita,fra, spa,hun,afr,msa,tgl,jav,ben, mar,est,hin,urd,fin} | 100k | | ✓ |
| TYDI-L | ours: focus on TyDi-QA languages | eng–{ara,ben,swa,kor} | 80–469k | ✓ | ✓ |
| TYDI-M | | | 60k | ✓ | ✓ |
| TYDI-S | | | 10k | ✓ | ✓ |
| 111-S | ours: using 111 eng-X corpora | See Appendix B | max 1k | ✓ | ✓ |
| 111-L | | | max 5k | ✓ | ✓ |

Table 1: Details on the multilingual aligned models used in this study.

downstream task (Arabic, Bengali, Swahili, and Korean) to finetune both an mBERT and an XML-R model.

- TYDI-M: same as above, but we only use 60k parallel sentences for each language pair.
- TYDI-S: same as above, but we only use 10k parallel sentences for each language pair.
- 111-L: we use up to 5k parallel sentences between English and 111 languages to finetune both an mBERT and an XML-R model.
- 111-s: same as above, but we use a maximum of 1k parallel sentences per language pair.

**QA model fine-tuning** After the first step, we perform "task-tuning," training models on the two datasets (TyDi-QA-mono and TyDi-QA+aug) with the encoder initialized with the alignment-based fine-tuned checkpoints.

## 5 Results and evaluation

### 5.1 Evaluation on Synthetic Dataset

In Tables 2 and 3, we report the TyDi-QA primary task results (passage selection and minimal answer) on our cross-lingual custom test sets (see Appendix A for similarly detailed results on the development set). TyDi-QA uses F1 score as evaluation matric. Here we report (question, context) language-pairwise scores. The main takeaway is that both data augmentation and alignment-based fine-tuning lead to general improvements. We delve deeper into our analyses below.

**Effect of cross-lingual training dataset** In all cases, we observe that using the synthetic cross-lingual dataset for training (i.e. TyDi-QA+aug) with questions in a different language than the context significantly improves the performance on the cross-lingual QA. For the passage selection task the improvement is between 2 and 10 percentage points on average for either mBERT and XLM-R, while the gains are on average larger for the minimal answer task, ranging between 6 and 15 percentage points on average. Since the benefit of using these training data is clear, we use it for all following experiments.

**mBERT vs XLM-R** Our initial hypothesis was that XLM-R would be in general better than mBERT, since it (a) has been trained on more data, and (b) it was pre-trained for longer using the settings of the more-robust RoBERTa (Liu et al., 2019). However, our expectation is not confirmed by the results.

Even without any alignment-based fine-tuning, the performance of the initial checkpoints of the models when task-tuned monolingually is largely comparable. When the context is non-English $(Q_{en}, C_{xx})$, XLM-R is slightly better (c.f. average performances of 54.4 (mBERT) and 55.5 (XLM-R) for passage selection and 45.6 to 42.8 for the minimal answer task). When the context is English and the question is in another language, though, mBERT performs better (c.f. 48.0 to 45.5 and 28.9 to 25.6 for the two tasks).

However, XLM-R benefits a lot more from task-tuning with the synthetically augmented data (TyDi-QA-+AUG), in some cases improving by up to 15 percentage points on average (e.g. in minimal answer for $(Q_{xx}, C_{en})$, while mBERT only improves by up to 8 percentage points. As a result, the best performing baselines without alignment fine-tuning are XLM-R models.

Following alignment-based fine-tuning and task-training XLM-R generally yields better performance in terms of F-score. The only exceptions are in the case of Swahili context ($Q_{en}, C_{sw}$) for both tasks, where mBERT leads to higher F-score, with XLM-R lagging behind by a couple of percentage points.

| Model | Tr. Data | $Q_{en}, C_{ar}$ | $Q_{en}, C_{bn}$ | $Q_{en}, C_{sw}$ | $Q_{en}, C_{ko}$ | avg | $Q_{ar}, C_{en}$ | $Q_{bn}, C_{en}$ | $Q_{sw}, C_{en}$ | $Q_{ko}, C_{en}$ | avg |
|---|---|---|---|---|---|---|---|---|---|---|---|
| **Baselines:** | | | | | | | | | | | |
| mBERT | mono | 76.4 | 43.0 | 60.9 | 37.3 | 54.4 | 51.8 | 43.3 | 46.9 | 49.8 | 48.0 |
| mBERT | +aug | 80.3 | 51.1 | 64.1 | 47.6 | 60.8 | 51.8 | 49.6 | 52.1 | 50.0 | 50.9 |
| XLM-R | mono | 78.3 | 46.6 | 61.8 | 35.3 | 55.5 | 47.6 | 44.6 | 38.7 | 51.3 | 45.5 |
| XLM-R | +aug | 81.1 | 56.4 | 64.3 | 50.1 | **63.0** | 56.1 | 55.0 | 55.3 | 56.6 | 55.7 |
| **mBERT +Alignment FT**: | | | | | | | | | | | |
| +cao-high | +aug | 78.6 (✗) | 49.4 (✗) | 61.2 (✗) | 44.6 (✗) | 58.4 | 51.4 (✗) | 49.6 (✗) | 51.7 (✗) | 51.1 (✗) | 50.9 |
| +zhao-low | +aug | 79.2 (✗) | 53.4 (✓) | 62.1 (✗) | 48.5 (✗) | 60.8 | 53.2 (✗) | 48.1 (✓) | 51.4 (✗) | 51.4 (✗) | 51.0 |
| +tydi-l | +aug | 81.2 (✓) | 52.1 (✓) | **65.2** (✓)* | 49.8 (✓) | 62.1 | 53.2 (✓) | 52.5 (✓) | 53.5 (✓) | 52.9 (✓) | 53.0 |
| +111-s | +aug | 79.4 (✓) | 54.7 (✓) | 64.2 (✓) | 47.9 (✓) | 61.6 | 52.5 (✓) | 50.0 (✓) | 50.5 (✓) | 51.7 (✓) | 51.2 |
| +111-l | +aug | 80.3 (✓) | 49.6 (✓) | 63.9 (✓) | 47.2 (✓) | 60.2 | 52.8 (✓) | 48.4 (✓) | 49.2 (✓) | 48.7 (✓) | 49.8 |
| **XLM-R +Alignment FT**: | | | | | | | | | | | |
| +zhao-(low+high) | +aug | 80.7 (✗) | 51.0 (✓) | 63.8 (✗) | 47.9 (✗) | 60.9 | 57.0 (✗) | 55.5 (✓) | 55.1 (✗) | 56.1 (✗) | 55.9 |
| +tydi-l | +aug | 81.0 (✓) | 50.5 (✓) | 62.2 (✓) | 50.2 (✓) | 61.0 | **57.7** (✓)* | 55.7 (✓) | **55.9** (✓)* | 56.5 (✓) | **56.5** |
| +111-s | +aug | 80.4 (✓) | **56.8** (✓)* | 61.7 (✓) | 50.5 (✓) | 62.3 | 57.3 (✓) | 55.4 (✓) | 54.9 (✓) | **56.7** (✓) | 56.1 |
| +111-l | +aug | **81.3** (✓) | 52.7 (✓) | 62.4 (✓) | **51.5** (✓)* | 62.0 | 57.6 (✓) | **56.1** (✓)* | 54.4 (✓) | 56.0 (✓) | 56.0 |

Table 2: Cross-lingual passage selection results (test set). ✓/✗: the language is/isn't included in the finetuning mix. *:denotes statistically significant better system than the corresponding baseline with $p < 0.05$.

| Model | Tr. Data | $Q_{en}, C_{ar}$ | $Q_{en}, C_{bn}$ | $Q_{en}, C_{sw}$ | $Q_{en}, C_{ko}$ | avg | $Q_{ar}, C_{en}$ | $Q_{bn}, C_{en}$ | $Q_{sw}, C_{en}$ | $Q_{ko}, C_{en}$ | avg |
|---|---|---|---|---|---|---|---|---|---|---|---|
| **Baselines:** | | | | | | | | | | | |
| mBERT | mono | 61.3 | 32.0 | 50.3 | 27.4 | 42.8 | 38.0 | 20.9 | 25.2 | 31.5 | 28.9 |
| mBERT | +aug | 68.9 | 42.4 | 53.4 | 35.7 | 50.1 | 36.4 | 33.0 | 33.9 | 34.7 | 34.5 |
| XLM-R | mono | 65.8 | 36.3 | 51.9 | 28.6 | 45.6 | 29.2 | 26.2 | 15.2 | 31.8 | 25.6 |
| XLM-R | +aug | 70.0 | 47.6 | 55.9 | 37.5 | **52.8** | 41.8 | 38.8 | 40.0 | 42.0 | 40.6 |
| **mBERT +Alignment FT**: | | | | | | | | | | | |
| +cao-high | +aug | 66.1 (✗) | 40.3 (✗) | 52.4 (✗) | 31.0 (✗) | 47.4 | 35.5 (✗) | 32.2 (✗) | 33.0 (✗) | 33.8 (✗) | 33.6 |
| +zhao-low | +aug | 68.5 (✗) | 39.0 (✓) | 53.1 (✗) | 33.6 (✗) | 48.5 | 36.6 (✗) | 32.1 (✓) | 32.3 (✗) | 34.5 (✗) | 33.9 |
| +tydi-l | +aug | 68.9 (✓) | 45.8 (✓) | **56.4** (✓)* | 37.0 (✓) | 52.0 | 36.5 (✓) | 32.0 (✓) | 35.1 (✓) | 35.1 (✓) | 34.7 |
| +111-s | +aug | 69.1 (✓) | 42.5 (✓) | 54.4 (✓) | 36.2 (✓) | 50.5 | 37.0 (✓) | 33.5 (✓) | 34.4 (✓) | 36.2 (✓) | 35.3 |
| +111-l | +aug | 68.6 (✓) | 41.0 (✓) | 52.7 (✓) | 35.2 (✓) | 49.4 | 36.7 (✓) | 32.8 (✓) | 32.3 (✓) | 32.6 (✓) | 33.6 |
| **XLM-R +Alignment FT**: | | | | | | | | | | | |
| +zhao-(low+high) | +aug | 68.3 (✗) | 42.0 (✓) | 54.0 (✗) | 39.0 (✗) | 50.8 | 40.9 (✗) | 39.4 (✓) | 38.6 (✗) | 40.4 (✗) | 39.8 |
| +tydi-l | +aug | **70.4** (✓) | 45.6 (✓) | 55.2 (✓) | 38.4 (✓) | 52.4 | **43.3** (✓)* | **41.0** (✓)* | **42.2** (✓)* | **42.3** (✓)* | **42.2** |
| +111-s | +aug | 69.6 (✓) | **48.9** (✓) | 53.0 (✓) | **39.4** (✓)* | 52.7 | 42.8 (✓) | 40.5 (✓) | 41.1 (✓) | 41.5 (✓) | 41.5 |
| +111-l | +aug | 69.5 (✓) | 44.6 (✓) | 54.9 (✓) | 39.2 (✓) | 52.0 | 42.4 (✓) | 39.6 (✓) | 40.7 (✓) | 41.8 (✓) | 41.1 |

Table 3: Cross-lingual minimal answer results (test set). ✓/✗: the language is/isn't included in the finetuning mix. *:denotes statistically significant better system than the corresponding baseline with $p < 0.05$.

**Effect of language choices in multilingual alignment** The comparison of different multilingually-aligned checkpoints leads to two main findings. First, that including the evaluation language in the alignment fine-tuning state is important in downstream performance. Across both tasks and all language settings, the highest performing setting is one where the language pair was included in the fine-tuning (marked with a checkmark ✓ in the two results Tables).

Aligning the representations of languages other than the ones we evaluate on does not seem to lead to improvements. The cao-high, zhao-low, and zhao-(low+high) models generally perform 1-2 percentage points worse than the comparable mBERT or XLM-R baselines using the same task-tuning data (+aug). This is an indication of negative interference (Wang et al., 2020), which we suspect is due to the models using a large amount of data in a limited set of languages that overfits the representations to these languages.

In contrast, our 111-s and 111-l models, despite including a lot more languages in the alignment fine-tuning stage with less data per language, suffer from less negative interference (as evidenced by performing on average just 1-2 percentage points less than the TYDI-L model - the other models can be up to 5 points worse than TYDI-L). In fact, the 111-X XLM-R models are the best ones in some settings for Bengali and Korean (e.g. $Q_{en}, C_{bn}$ and $Q_{en}, C_{ko}$). This observation implies that such fine-tuning on as many languages as possible is perhaps a viable strategy towards improving downstream performance on as many languages as possible, and not just the four languages that we evaluate on. We believe that studying the phenomenon of negative interference and attempting to mitigate it is a very promising avenue for future work towards building robust, more equitable cross-lingual QA systems.

**The effect of data size in alignment fine-tuning** To quantify the effect of including more (or less) parallel data in the alignment fine-tuning stage, we perform an ablation study varying the training data size in the two language settings: (a) when we only use the TyDi-QA languages, varying the data to have up to 10k (TYDI-S), 60k (TYDI-M), or using all available parallel data with between 80-260k sentences per language (TYDI-L, and (b) when using 111 languages, we use up to 1k (111-S) or 5k (111-L) parallel sentences per language.

Results with the two 111-X models were already reported in Tables 2 and 3. Results with the TyDi-QA-focused ablations are listed in Table 4 for both tasks. When focusing on the few languages of TyDi-QA, we find that in most cases using more data leads to better overall performance, but restricting ourselves to even just 10k parallel sentences per language still leads to comparable results with similar improvements. In the case of passage selection with English questions ($Q_{en}, C_{xx}$), in fact, using less data leads to slightly higher average performance (c.f. 61.3 to 61.0). This is an encouraging result, since it reveals that large parallel datasets (which for most of the world's languages are not available anyway) are not a hard requirement.

When using a larger pool of languages, a 5-fold increase in data is not beneficial, with the performance of 111-L being worse than the 111-S overall. Beyond the potential negative interference effects discussed previously, we hypothesize that the increased amount of language data creates a data imbalance as there are languages with very little parallel data (i.e. around 500 sentences). In future work we will investigate whether data balancing schemes such as the one used by Siddhant et al. (2020) can mitigate this effect and potentially allow us to leverage all available data.

**Statistical significance** Beyond simply calculating evaluation scores (i.e. F-score), we conducted statistical significance tests comparing the baseline (TyDi-QA+`aug` data) to the two best performing models: 111-s and TYDI-L. We perform pairwise bootstrap re-sampling (Koehn, 2004) between the model predictions for both types of pretrained models (i.e. MBERT and XLM-R) and for both type of tasks (i.e. passage selection and minimal answer). While we don't observe statistically significant improvement in all results, in most cases our models perform better than the baseline model at a 95% confidence interval.

## 5.2 Human translations evaluation

Table 5 lists the results obtained on the real-world scenario of spoken and consequently transcribed questions. We remind the reader that we collected such data for two settings: $(Q_{en}, C_{bn})$ and $(Q_{en}, C_{sw})$, asking bilingual speakers native in Bengali and Swahili to translate questions in English, which we then transcribed using publicly available ASR systems, while also hand-creating gold transcriptions for Bengali.[9]

In this real world scenario, the results are more mixed. As before, we find that XLM-R generally performs better than mBERT, and that task-tuning with the synthetically augmented data helps significantly. For the passage selection task, interestingly, the alignment-finetuned models lag behind the best baseline for all settings. For the minimal answer task, the alignment-based fine-tuning does lead to additional improvements, increasing the obtained F-score by around 2 percentage points over the best baseline. In these cases, the best performing model is the one where we used data from 111 languages for alignment – in such noisy settings, as in the case of the two automatic transcriptions, using only the TyDi-QA languages is inferior. Of importance, also, is the general drop in performance when comparing the gold Bengali question transcriptions to the automatic ones, denoting that future work is required to make QA models robust to ASR noise, as Ravichander et al. (2021) and Faisal et al. (2021) have noted.

Last, we compare the above scenario with the scenario where the speakers asked the same questions in their native language. The results for this monolingual setting are at the bottom row of Table 5, and are comparable to the results in the rest of the Table (column-wise), as they are over the exact same test set. Notably, the monolingual scenario yields more than 10 percentage points improvements for passage selection in Bengali (and 2 points for the minimal answer task), but it is much worse, by more than 15 percentage points for both tasks in Swahili. This is due to the poor quality of the ASR transcription for Swahili, as Faisal et al. (2021) discuss. This means that a Swahili speaker who also speaks English would receive almost 60% more utility out of our systems if they also speak and can ask their question in English. This highlights the need for advancing both the monolingual *and* cross-lingual capabilities of QA systems, as well as the need for making these systems robust to noise and other variations. Furthermore, we emphasize the need for realistic datasets that reflect the users usage of QA systems across more of the world's languages and language varieties.

[9] One of the authors is a native Bengali speaker.

| Model | Tr. Data | $Q_{en}, C_{ar}$ | $Q_{en}, C_{bn}$ | $Q_{en}, C_{sw}$ | $Q_{en}, C_{ko}$ | avg | $Q_{ar}, C_{en}$ | $Q_{bn}, C_{en}$ | $Q_{sw}, C_{en}$ | $Q_{ko}, C_{en}$ | avg |
|---|---|---|---|---|---|---|---|---|---|---|---|
| **Passage selection**: | | | | | | | | | | | |
| +TYDI-S | | 80.9 | 51.7 | 61.8 | **50.7** | **61.3** | 57.0 | **56.5** | 55.4 | 54.8 | 55.9 |
| +TYDI-M | +aug | 81.0 | **51.7** | 61.1 | 50.4 | 61.0 | 57.6 | 54.9 | 53.7 | 56.3 | 55.6 |
| +TYDI-L | | **81.0** | 50.5 | **62.2** | 50.2 | 61.0 | **57.7** | 55.7 | **55.9** | **56.5** | **56.5** |
| **Minimal answer**: | | | | | | | | | | | |
| +TYDI-S | | **71.2** | 44.6 | 55.1 | 37.7 | 52.2 | 42.8 | 38.7 | 40.0 | 41.5 | 40.8 |
| +TYDI-M | +aug | 71.1 | **46.5** | 54.1 | 36.8 | 52.1 | 42.8 | 40.7 | 39.0 | 41.1 | 40.9 |
| +TYDI-L | | 70.4 | 45.6 | **55.2** | **38.4** | **52.4** | **43.3** | **41.0** | **42.2** | **42.3** | **42.2** |

Table 4: Primary Task Results (XLM-R) varying training size for alignment.

We further analyze the cross-lingual results in Table 6, where we compare the correct-incorrect frequency of minimal answers for Bengali and Swahili in two settings: asking questions in the context language or in the English translations (human-recorded, asr transcriptions). Here we only report comparison on the answers which are either fully correct or incorrect leaving the partially correct ones. Overall, our models get less instances completely wrong. Interestingly, we observe that in a number of cases, alignment based fine-tuning helps in predicting correct answers in cross lingual setting which were incorrectly predicted in a monolingual setting: for example, in XLM-R for Swahili, 101 instances were originally wrong regardless of the language of the question (English or Swahili). Our version of XLM-R, though, gets less examples completely wrong (93 vs 101) and gets the answer correct in at least one of the settings. This categorization of the dev/test instances could perhaps further help classify examples into easy or hard for multilingual models and be of further use in further studies of multilingual fairness and robustness.

## 6 Related Work

A significant amount of work is devoted to studying and improving the cross-lingual capabilities of QA models. TyDi-QA (Clark et al., 2020) is a notable recent dataset focusing on the inclusion of 11 typologically diverse languages. XOR-QA (Asai et al., 2021) builds on top of TyDi-QA, exploring open domain QA systems, where the search for an answer to a question unanswerable in the original language integrates translated resources from relevant English Wikipedia pages, in a task reflective of our setting. We view our work as orthogonal to XOR-QA, as Asai et al. (2021) put more emphasis on the retrieval of relevant passages rather than the cross-lingual capabilities of the QA ("reader") model per se.

There exist a number of multilingual QA benchmarks, including MLQA (Lewis et al., 2020), MKQA (Longpre et al., 2020) and XQuAD (Artetxe et al., 2020). MLQA translates original English questions to 7 other languages to train a multilingual QA model. MKQA comprises of questions from 26 diverse languages. XQuAD uses translated questions from SQuAD (Rajpurkar et al., 2016) (originally in English) to prepare a widely used and easily adaptable multilingual benchmark of SQuAD baseline. In SD-QA (Faisal et al., 2021) from which our work is inspired, the authors prepare a naturally spoken version of TyDi-QA over 5 languages. In this work, we further expand the scope to study the cross-lingual abilities of QA systems for these 5 languages.

Using cross-lingual objectives that leverage parallel language data is a promising direction towards improving the cross-lingual abilities of lan-guage models (Conneau and Lample, 2019). Cao et al. (2020) fine-tuned mBERT on a parallel corpus (taken from Europarl) using word-level alignments obtained with fast-align (Dyer et al., 2013). This model aims to decrease the representation distance between words with similar meanings across languages. Zhao et al. (2021) used the fine-tuning process defined by Cao et al. (2020) and further tuned it for low resource languages. After the alignment fine-tuning stage, the authors also perform last-layer embedding normalization and language specific word-word coordination to further improve on downstream tasks. These contextual representation alignment works are evaluated on tasks designed for cross-lingual and zero-shot transfer like XNLI (Conneau et al., 2018) or RFEVAL (Zhao et al., 2020). Our work is the first to evaluate these methods on the QA task, but also the first to expand the alignment-based fine-tuning to include almost all languages used in the pre-trained models, as opposed to only using the languages on which evaluation is performed.

| Model | Tr. Data | Passage Selection $Q_{en}, C_{bn}$ (Gold Transc.) | Passage Selection $Q_{en}, C_{bn}$ (ASR) | Passage Selection $Q_{en}, C_{sw}$ (ASR) | Minimal Answer $Q_{en}, C_{bn}$ (Gold Transc.) | Minimal Answer $Q_{en}, C_{bn}$ (ASR) | Minimal Answer $Q_{en}, C_{sw}$ (ASR) |
|---|---|---|---|---|---|---|---|
| **Baselines:** | | | | | | | |
| mBERT | mono | 41.1 | 37.2 | 53.4 | 32.0 | 26.5 | 43.1 |
| mBERT | +aug | 46.6 | 45.9 | 54.1 | 38.8 | 36.4 | 45.0 |
| XLM-R | mono | 41.0 | 37.0 | 52.6 | 28.7 | 25.1 | 42.1 |
| XLM-R | +aug | **55.2** | **52.4** | **59.0** | 43.8 | 41.6 | 47.4 |
| **mBERT +Alignment FT**: | | | | | | | |
| +CAO-HIGH | | 48.4 (✗) | 40.7 (✗) | 53.2 (✗) | 38.0 (✗) | 31.5 (✗) | 42.6 (✗) |
| +ZHAO-LOW | | 47.4 (✓) | 42.1 (✓) | 51.6 (✗) | 38.6 (✓) | 32.4 (✓) | 42.3 (✗) |
| +TYDI-L | +aug | 45.9 (✓) | 39.6 (✓) | 54.0 (✓) | 39.1 (✓) | 33.7 (✓) | 44.2 (✓) |
| +111-S | | 51.0 (✓) | 45.1 (✓) | 51.4 (✓) | 38.6 (✓) | 34.0 (✓) | 41.7 (✓) |
| +111-L | | 47.5 (✓) | 43.5 (✓) | 52.3 (✓) | 37.7 (✓) | 36.1 (✓) | 41.8 (✓) |
| **XLM-R +Alignment FT**: | | | | | | | |
| +ZHAO-(LOW+HIGH) | | 52.5 (✓) | 50.8 (✓) | 56.9 (✗) | 43.4 (✓) | 39.8 (✓) | 47.1 (✗) |
| +TYDI-L | +aug | 49.3 (✓) | 47.5 (✓) | 57.0 (✓) | 40.5 (✓) | 38.7 (✓) | 48.2 (✓) |
| +111-S | | 52.8 (✓) | 50.8 (✓) | 58.6 (✓) | **45.6** (✓) | **43.4** (✓) | 48.2 (✓) |
| +111-L | | 49.6 (✓) | 48.9 (✓) | 57.1 (✓) | 41.7 (✓) | 41.0 (✓) | **49.2** (✓) |
| **Monolingual Setting:** **(no translation)** | | $Q_{bn}, C_{bn}$ 61.9 | $Q_{bn}, C_{bn}$ 60.3 | $Q_{sw}, C_{sw}$ 43.8 | $Q_{bn}, C_{bn}$ 47.9 | $Q_{bn}, C_{bn}$ 47.2 | $Q_{sw}, C_{sw}$ 31.4 |

Table 5: Primary tasks result for the real-world scenario (human-translated and ASR-transcribed English questions) over foreign contexts (test set).

| | | Baselines mBERT | | Baselines XLM-R | | mBERT+Align. FT 111-S | | XLM-R+Align. FT 111-S | |
|---|---|---|---|---|---|---|---|---|---|
| | | $Q_{en}, C_{bn}$ | | $Q_{en}, C_{bn}$ | | $Q_{en}, C_{bn}$ | | $Q_{en}, C_{bn}$ | |
| | | Correct | Wrong | Correct | Wrong | Correct | Wrong | Correct | Wrong |
| $Q_{bn}, C_{bn}$ | Correct | 34 | 14 | **43** | 12 | 32 | 14 | 42 | **8** |
| | Wrong | 6 | 41 | 2 | **33** | 3 | 42 | **6** | 34 |
| | | $Q_{en}, C_{sw}$ | | $Q_{en}, C_{sw}$ | | $Q_{en}, C_{sw}$ | | $Q_{en}, C_{sw}$ | |
| | | Correct | Wrong | Correct | Wrong | Correct | Wrong | Correct | Wrong |
| $Q_{sw}, C_{sw}$ | Correct | 222 | 27 | **229** | 20 | 228 | 32 | 221 | **18** |
| | Wrong | 10 | 108 | 8 | 101 | 12 | 115 | **15** | **93** |

Table 6: Frequency comparison of correct/incorrect min-F1 using original and translated questions. The baseline ones are trained on augmented data and evaluated on ASR outputs of human translations.

## 7 Conclusion and Future Work

In this work, we have studied the cross-lingual extractive QA setting where the question and context-to-search are in different languages. Through experiments on synthetic and newly collected data in 4 languages, we have shown that data augmentation along with alignment-based fine-tuning can lead to big improvements in downstream performance. In future work, we plan to collect a larger dataset covering more languages for such cross-lingual settings. We also aim to extend this study using even more parallel data whenever available, as well as to investigate the feasibility of using language-specific tools in parts of the system's pipelines (e.g. in segmentation or tokenization).

## Acknowledgments

This work is generously supported by NSF Awards IIS-2040926 and IIS-2125466. The dataset creation was supported though a Google Award for Inclusion Research. We also want to thank Jacob Eisenstein, Manaal Faruqui, and Jon Clark for helpful discussions on question answering and data collection. The authors are grateful to Kathleen Siminyu for her help with collecting Kiswahili and Kenyan English speech samples.

## References

Chris Alberti, Kenton Lee, and Michael Collins. 2019. A BERT baseline for the natural questions.

Antonios Anastasopoulos and Graham Neubig. 2020. Should all cross-lingual embeddings speak english? In *Proceedings of the 58th Annual Meeting of the Association for Computational Linguistics*, pages 8658–8679, Online. Association for Computational Linguistics.

Mikel Artetxe, Sebastian Ruder, and Dani Yogatama. 2020. On the cross-lingual transferability of monolingual representations. In *Proceedings of the 58th Annual Meeting of the Association for Computational Linguistics*, pages 4623–4637, Online. Association for Computational Linguistics.

Akari Asai, Jungo Kasai, Jonathan Clark, Kenton Lee, Eunsol Choi, and Hannaneh Hajishirzi. 2021. XOR QA: Cross-lingual open-retrieval question answering. In *Proceedings of the 2021 Conference of the North American Chapter of the Association for Computational Linguistics: Human Language Technologies*, pages 547–564, Online. Association for Computational Linguistics.

Steven Cao, Nikita Kitaev, and Dan Klein. 2020. Multilingual alignment of contextual word representations. In *International Conference on Learning Representations*.

Jonathan H Clark, Eunsol Choi, Michael Collins, Dan Garrette, Tom Kwiatkowski, Vitaly Nikolaev, and Jennimaria Palomaki. 2020. TyDi QA: A benchmark for information-seeking question answering in typologically diverse languages. *Transactions of the Association for Computational Linguistics*, 8:454–470.

Alexis Conneau, Kartikay Khandelwal, Naman Goyal, Vishrav Chaudhary, Guillaume Wenzek, Francisco Guzmán, Edouard Grave, Myle Ott, Luke Zettlemoyer, and Veselin Stoyanov. 2020. Unsupervised cross-lingual representation learning at scale. In *Proceedings of the 58th Annual Meeting of the Association for Computational Linguistics*, pages 8440–8451, Online. Association for Computational Linguistics.

Alexis Conneau and Guillaume Lample. 2019. Cross-lingual language model pretraining. *Advances in Neural Information Processing Systems*, 32:7059–7069.

Alexis Conneau, Ruty Rinott, Guillaume Lample, Adina Williams, Samuel R. Bowman, Holger Schwenk, and Veselin Stoyanov. 2018. Xnli: Evaluating cross-lingual sentence representations. In *Proceedings of the 2018 Conference on Empirical Methods in Natural Language Processing*. Association for Computational Linguistics.

Arnab Debnath, Navid Rajabi, Fardina Fathmiul Alam, and Antonios Anastasopoulos. 2021. Towards more equitable question answering systems: How much more data do you need? In *Proceedings of the 59th Annual Meeting of the Association for Computational Linguistics and the 11th International Joint Conference on Natural Language Processing (ACL-IJCNLP)*. Association for Computational Linguistics.

Jacob Devlin, Ming-Wei Chang, Kenton Lee, and Kristina Toutanova. 2019. Bert: Pre-training of deep bidirectional transformers for language understanding. In *Proceedings of the 2019 Conference of the North American Chapter of the Association for Computational Linguistics: Human Language Technologies, Volume 1 (Long and Short Papers)*, pages 4171–4186.

Zi-Yi Dou and Graham Neubig. 2021. Word alignment by fine-tuning embeddings on parallel corpora. In *Conference of the European Chapter of the Association for Computational Linguistics (EACL)*.

Chris Dyer, Victor Chahuneau, and Noah A. Smith. 2013. A simple, fast, and effective reparameterization of IBM model 2. In *Proceedings of the 2013 Conference of the North American Chapter of the Association for Computational Linguistics: Human Language Technologies*, pages 644–648, Atlanta, Georgia. Association for Computational Linguistics.

Ahmed El-Kishky, Vishrav Chaudhary, Francisco Guzmán, and Philipp Koehn. 2020. CCAligned: A massive collection of cross-lingual web-document pairs. In *Proceedings of the 2020 Conference on Empirical Methods in Natural Language Processing (EMNLP)*, pages 5960–5969, Online. Association for Computational Linguistics.

Fahim Faisal, Sharlina Keshava, Md Mahfuz ibn Alam, and Antonios Anastasopoulos. 2021. SD-QA: Spoken Dialectal Question Answering for the Real World. In *Findings of the 2021 Conference on Empirical Methods in Natural Language Processing (EMNLP Findings)*. Association for Computational Linguistics.

Manaal Faruqui, Jesse Dodge, Sujay Kumar Jauhar, Chris Dyer, Eduard Hovy, and Noah A. Smith. 2015. Retrofitting word vectors to semantic lexicons. In *Proceedings of the 2015 Conference of the North American Chapter of the Association for Computational Linguistics: Human Language Technologies*, pages 1606–1615, Denver, Colorado. Association for Computational Linguistics.

Robert M. French. 1999. Catastrophic forgetting in connectionist networks. *Trends in Cognitive Sciences*, 3(4):128–135.

Jianfeng Gao, Michel Galley, and Lihong Li. 2018. Neural approaches to conversational ai. *The 41st International ACM SIGIR Conference on Research & Development in Information Retrieval*.

Kuan-Hao Huang, Wasi Uddin Ahmad, Nanyun Peng, and Kai-Wei Chang. 2021. Improving zero-shot cross-lingual transfer learning via robust training.

Zhengbao Jiang, Antonios Anastasopoulos, Jun Araki, Haibo Ding, and Graham Neubig. 2020. X-factr: Multilingual factual knowledge retrieval from pretrained language models. In *Proceedings of the 2020 Conference on Empirical Methods in Natural Language Processing (EMNLP)*, pages 5943–5959, Online. Association for Computational Linguistics.

Philipp Koehn. 2004. Statistical significance tests for machine translation evaluation. In *Proceedings of the 2004 Conference on Empirical Methods in Natural Language Processing*, pages 388–395, Barcelona, Spain. Association for Computational Linguistics.

Saurabh Kulshreshtha, Jose Luis Redondo Garcia, and Ching Yun Chang. 2020. Cross-lingual alignment methods for multilingual bert: A comparative study. In *Proceedings of the 2020 Conference on Empirical Methods in Natural Language Processing: Findings*, pages 933–942.

Tom Kwiatkowski, Jennimaria Palomaki, Olivia Redfield, Michael Collins, Ankur Parikh, Chris Alberti, Danielle Epstein, Illia Polosukhin, Matthew Kelcey, Jacob Devlin, Kenton Lee, Kristina N. Toutanova, Llion Jones, Ming-Wei Chang, Andrew Dai, Jakob Uszkoreit, Quoc Le, and Slav Petrov. 2019. Natural questions: a benchmark for question answering research. *Transactions of the Association of Computational Linguistics*.

Patrick Lewis, Barlas Oguz, Ruty Rinott, Sebastian Riedel, and Holger Schwenk. 2020. Mlqa: Evaluating cross-lingual extractive question answering. In *Proceedings of the 58th Annual Meeting of the Association for Computational Linguistics*, pages 7315–7330.

Yinhan Liu, Myle Ott, Naman Goyal, Jingfei Du, Mandar Joshi, Danqi Chen, Omer Levy, Mike Lewis, Luke Zettlemoyer, and Veselin Stoyanov. 2019. Roberta: A robustly optimized bert pretraining approach. arXiv:1907.11692.

Shayne Longpre, Yi Lu, and Joachim Daiber. 2020. MKQA: A linguistically diverse benchmark for multilingual open domain question answering.

Michael McCloskey and Neal J. Cohen. 1989. Catastrophic interference in connectionist networks: The sequential learning problem. volume 24 of *Psychology of Learning and Motivation*, pages 109–165. Academic Press.

Benjamin Muller, Antonios Anastasopoulos, Benoît Sagot, and Djamé Seddah. 2020. When being unseen from mBERT is just the beginning: Handling new languages with multilingual LMs. In *Proceedings of the 2021 Conference of the North American Chapter of the Association for Computational Linguistics: Human Language Technologies*, Online. Association for Computational Linguistics.

Pranav Rajpurkar, Jian Zhang, Konstantin Lopyrev, and Percy Liang. 2016. SQuAD: 100,000+ questions for machine comprehension of text. In *Proceedings of the 2016 Conference on Empirical Methods in Natural Language Processing*, pages 2383–2392, Austin, Texas. Association for Computational Linguistics.

Roger Ratcliff. 1990. Connectionist models of recognition memory: Constraints imposed by learning and forgetting functions. *Psychological Review*, 97(2):285–308.

Abhilasha Ravichander, Siddharth Dalmia, Maria Ryskina, Florian Metze, Eduard Hovy, and Alan W Black. 2021. NoiseQA: Challenge set evaluation for user-centric question answering. In *Proceedings of the 16th Conference of the European Chapter of the Association for Computational Linguistics: Main Volume*, pages 2976–2992, Online. Association for Computational Linguistics.

Anna Rogers, Matt Gardner, and Isabelle Augenstein. 2021. Qa dataset explosion: A taxonomy of nlp resources for question answering and reading comprehension. arXiv:2107.12708.

Holger Schwenk, Vishrav Chaudhary, Shuo Sun, Hongyu Gong, and Francisco Guzmán. 2021. WikiMatrix: Mining 135M parallel sentences in 1620 language pairs from Wikipedia. In *Proceedings of the 16th Conference of the European Chapter of the Association for Computational Linguistics: Main Volume*, pages 1351–1361, Online. Association for Computational Linguistics.

Aditya Siddhant, Melvin Johnson, Henry Tsai, Naveen Ari, Jason Riesa, Ankur Bapna, Orhan Firat, and Karthik Raman. 2020. Evaluating the cross-lingual effectiveness of massively multilingual neural machine translation. In *Proceedings of the AAAI Conference on Artificial Intelligence*, volume 34, pages 8854–8861.

Jörg Tiedemann. 2012. Parallel data, tools and interfaces in OPUS. In *Proceedings of the Eight International Conference on Language Resources and Evaluation (LREC'12)*, Istanbul, Turkey. European Language Resources Association (ELRA).

Zirui Wang, Zachary C. Lipton, and Yulia Tsvetkov. 2020. On negative interference in multilingual models: Findings and a meta-learning treatment. In *Proceedings of the 2020 Conference on Empirical Methods in Natural Language Processing (EMNLP)*, pages 4438–4450, Online. Association for Computational Linguistics.

Yi Yang, Wen-tau Yih, and Christopher Meek. 2015. WikiQA: A challenge dataset for open-domain question answering. In *Proceedings of the 2015 Conference on Empirical Methods in Natural Language Processing*, pages 2013–2018, Lisbon, Portugal. Association for Computational Linguistics.

Wei Zhao, Steffen Eger, Johannes Bjerva, and Isabelle Augenstein. 2021. Inducing language-agnostic multilingual representations. In *Proceedings of *SEM 2021: The Tenth Joint Conference on Lexical and Computational Semantics*, pages 229–240, Online. Association for Computational Linguistics.

Wei Zhao, Goran Glavaš, Maxime Peyrard, Yang Gao, Robert West, and Steffen Eger. 2020. On the limitations of cross-lingual encoders as exposed by reference-free machine translation evaluation. In *Proceedings of the 58th Annual Meeting of the Association for Computational Linguistics*, pages 1656–1671, Online. Association for Computational Linguistics.

## A Evaluation on custom development set

In this section, we report the experimental result on TyDi-QA custom development set. See Table 7 for passage selection and Table 8 for minimal answer results

## B Parallel corpus selection

In Table 9, we present the parallel language statistics used in our multilingual alignment fine-tuning for the 111-s and 111-L models. We used in total 111 X-English parallel language datasets. Among these languages, 86 were used to pre-train mBERT and 87 were used during XLM-R pre-training. For training model 111-s, the maximum number of sentences per language pair was set to 1000, whereas we set the threshold to 5000 for model 111-L. We used parallel data from OPUS-100 (Tiedemann, 2012), WikiMatrix (Schwenk et al., 2021) and CCAligned (El-Kishky et al., 2020).

| Model | Tr. Data | $Q_{en}, C_{ar}$ | $Q_{en}, C_{bn}$ | $Q_{en}, C_{sw}$ | $Q_{en}, C_{ko}$ | avg |
|---|---|---|---|---|---|---|
| **Best Baselines:** | | | | | | |
| mBERT | mono | 60.2 | 41.9 | 61.8 | 36.5 | 50.1 |
| mBERT | +aug | 65.6 | 51.9 | 68.6 | 37.3 | 55.8 |
| XLM-R | mono | 58.7 | 29.7 | 62.5 | 25.4 | 44.1 |
| XLM-R | +aug | 65.1 | 51.9 | 69.0 | 41.7 | 56.9 |
| **mBERT +Alignment FT**: | | | | | | |
| +CAO-HIGH | +aug | 63.9 (✗) | 45.3 (✗) | 66.2 (✗) | 34.3 (✗) | 52.4 |
| +ZHAO-LOW | +aug | 67.2 (✗) | 46.4 (✓) | 67.2 (✗) | 45.1 (✗) | 56.5 |
| +TYDI-L | +aug | 64.7 (✓) | 48.1 (✓) | **69.9** (✓) | 43.8 (✓) | 56.6 |
| +111-S | +aug | 66.0 (✓) | 47.3 (✓) | 68.6 (✓) | **44.1** (✓) | 56.5 |
| +111-L | +aug | **66.8** (✓) | 44.4 (✓) | 68.3 (✓) | 41.1 (✓) | 55.1 |
| **XLM-R +Alignment FT**: | | | | | | |
| +ZHAO-(LOW+HIGH) | +aug | 64.3 (✗) | 48.1 (✓) | 69.2 (✗) | 32.3 (✗) | 53.5 |
| +TYDI-L | +aug | 64.3 (✓) | **52.5** (✓) | 69.4 (✓) | 39.5 (✓) | 56.4 |
| +111-S | +aug | 64.4 (✓) | 49.4 (✓) | 68.8 (✓) | 39.0 (✓) | 55.4 |
| +111-L | +aug | 64.7 (✓) | 50.0 (✓) | 68.2 (✓) | 39.4 (✓) | 55.6 |

| Model | Tr. Data | $Q_{ar}, C_{en}$ | $Q_{bn}, C_{en}$ | $Q_{sw}, C_{en}$ | $Q_{ko}, C_{en}$ | avg |
|---|---|---|---|---|---|---|
| **Best Baselines:** | | | | | | |
| mBERT | mono | 49.2 | 41.0 | 43.1 | 47.2 | 45.1 |
| mBERT | +aug | 50.0 | 46.3 | 48.2 | 48.2 | 48.2 |
| XLM-R | mono | 45.8 | 43.3 | 37.3 | 47.6 | 43.5 |
| XLM-R | +aug | 51.0 | 50.9 | 50.6 | 50.8 | 50.8 |
| **mBERT +Alignment FT**: | | | | | | |
| +CAO-HIGH | +aug | 47.8 (✗) | 47.6 (✗) | 46.9 (✗) | 48.6 (✗) | 47.7 |
| +ZHAO-LOW | +aug | 48.6 (✗) | 46.7 (✓) | 46.1 (✗) | 47.9 (✗) | 47.3 |
| +TYDI-L | +aug | 48.9 (✓) | 46.4 (✓) | 47.1 (✓) | 47.0 (✓) | 47.4 |
| +111-S | +aug | 50.2 (✓) | 49.0 (✓) | 48.7 (✓) | 49.2 (✓) | 49.3 |
| +111-L | +aug | 48.3 (✓) | 47.0 (✓) | 46.2 (✓) | 48.3 (✓) | 47.5 |
| **XLM-R +Alignment FT**: | | | | | | |
| +ZHAO-(LOW+HIGH) | +aug | 52.1 (✗) | 50.6 (✓) | 50.2 (✗) | 50.5 (✗) | 50.9 |
| +TYDI-L | +aug | **53.4** (✓) | **53.5** (✓) | **52.3** (✓) | **53.2** (✓) | 53.1 |
| +111-S | +aug | 52.8 (✓) | 52.6 (✓) | 50.9 (✓) | 51.9 (✓) | 52.1 |
| +111-L | +aug | 51.4 (✓) | 51.2 (✓) | 49.1 (✓) | 51.6 (✓) | 50.8 |

Table 7: Passage Selection Results for Foreign questions over english contexts (dev set).✓/✗: the language is/isn't included in the finetuning mix.

| Model | Tr. Data | $Q_{en}, C_{ar}$ | $Q_{en}, C_{bn}$ | $Q_{en}, C_{sw}$ | $Q_{en}, C_{ko}$ | avg |
|---|---|---|---|---|---|---|
| **Best Baselines:** | | | | | | |
| mBERT | mono | 43.4 | 30.4 | 50.9 | 31.3 | 39.0 |
| mBERT | +aug | 50.4 | 43.5 | 55.9 | 29.7 | 44.9 |
| XLM-R | mono | 46.1 | 23.0 | 50.6 | 22.6 | 35.6 |
| XLM-R | +aug | 51.7 | 42.5 | 56.7 | 36.4 | 46.8 |
| **mBERT +Alignment FT**: | | | | | | |
| +CAO-HIGH | +aug | 48.6 (✗) | 36.0 (✗) | 54.4 (✗) | 23.0 (✗) | 40.5 |
| +ZHAO-LOW | +aug | 51.7 (✗) | 36.0 (✓) | 55.3 (✗) | 35.7 (✗) | 44.7 |
| +TYDI-L | +aug | 50.7 (✓) | 36.0 (✓) | **57.9** (✓) | **37.2** (✓) | 45.5 |
| +111-S | +aug | **52.2** (✓) | 35.9 (✓) | 56.6 (✓) | 32.9 (✓) | 44.4 |
| +111-L | +aug | 51.6 (✓) | 35.9 (✓) | 55.7 (✓) | 29.9 (✓) | 43.3 |
| **XLM-R +Alignment FT**: | | | | | | |
| +ZHAO-(LOW+HIGH) | +aug | 50.3 (✗) | 38.1 (✓) | 56.3 (✗) | 26.5 (✗) | 42.8 |
| +TYDI-L | +aug | 50.6 (✓) | **44.1** (✓) | 57.8 (✓) | 29.8 (✓) | 45.6 |
| +111-S | +aug | 49.2 (✓) | 37.4 (✓) | 56.7 (✓) | 29.5 (✓) | 43.2 |
| +111-L | +aug | 50.7 (✓) | 41.6 (✓) | 57.1 (✓) | 33.6 (✓) | 45.8 |

| Model | Tr. Data | $Q_{ar}, C_{en}$ | $Q_{bn}, C_{en}$ | $Q_{sw}, C_{en}$ | $Q_{ko}, C_{en}$ | avg |
|---|---|---|---|---|---|---|
| **Best Baselines:** | | | | | | |
| mBERT | mono | 33.2 | 17.9 | 22.5 | 29.2 | 25.7 |
| mBERT | +aug | 31.0 | 28.4 | 29.5 | 32.2 | 30.3 |
| XLM-R | mono | 26.1 | 22.6 | 14.7 | 29.3 | 23.2 |
| XLM-R | +aug | 34.4 | 34.4 | 35.1 | 34.5 | 34.6 |
| **mBERT +Alignment FT**: | | | | | | |
| +CAO-HIGH | +aug | 31.1 (✗) | 29.2 (✗) | 27.5 (✗) | 30.6 (✗) | 29.6 |
| +ZHAO-LOW | +aug | 32.0 (✗) | 28.4 (✓) | 29.1 (✗) | 30.9 (✗) | 30.1 |
| +TYDI-L | +aug | 29.8 (✓) | 28.8 (✓) | 28.2 (✓) | 31.3 (✓) | 29.5 |
| +111-S | +aug | 32.8 (✓) | 30.3 (✓) | 29.4 (✓) | 32.3 (✓) | 31.2 |
| +111-L | +aug | 29.6 (✓) | 28.1 (✓) | 28.0 (✓) | 29.4 (✓) | 28.8 |
| **XLM-R +Alignment FT**: | | | | | | |
| +ZHAO-(LOW+HIGH) | +aug | 34.8 (✗) | 33.3 (✓) | 33.8 (✗) | 34.8 (✗) | 34.2 |
| +TYDI-L | +aug | 36.4 (✓) | 34.8 (✓) | **35.4** (✓) | 36.6 (✓) | 35.8 |
| +111-S | +aug | 35.6 (✓) | 35.4 (✓) | 33.1 (✓) | 37.0 (✓) | 35.3 |
| +111-L | +aug | 37.7 (✓) | **36.2** (✓) | 35.2 (✓) | **37.9** (✓) | 36.8 |

Table 8: Minimal Answer Results for Foreign questions over english contexts (dev set).✓/✗: the language is/isn't included in the finetuning mix.

| Index | Language | Parallel language code | mBERT training | XLM-R training | Sentence count |
|---|---|---|---|---|---|
| 1 | afrikaans | af-en | | | 1507 |
| 2 | amharic | am-en | ✗ | | 4446 |
| 3 | aragonese | an-en | | ✗ | 463 |
| 4 | arabic | ar-en | | | 4174 |
| 5 | assamese | as-en | ✗ | | 5192 |
| 6 | azerbaijani | az-en | | | 1860 |
| 7 | bashkir | ba-en | | ✗ | 5753 |
| 8 | belarusian | be-en | | | 2556 |
| 9 | bulgarian | bg-en | | | 4131 |
| 10 | bengali | bn-en | | | 6873 |
| 11 | breton | br-en | | | 754 |
| 12 | bosnian | bs-en | | | 2017 |
| 13 | catalan | ca-en | | | 1563 |
| 14 | czech | cs-en | | | 2276 |
| 15 | chuvash | cv-en | | ✗ | 6299 |
| 16 | welsh | cy-en | | | 1513 |
| 17 | danish | da-en | | | 1858 |
| 18 | german | de-en | | | 1567 |
| 19 | dzongkha | dz-en | ✗ | ✗ | 215 |
| 20 | greek | el-en | | | 4514 |
| 21 | esperanto | eo-en | ✗ | | 1707 |
| 22 | spanish | es-en | | | 2405 |
| 23 | estonian | et-en | | | 2349 |
| 24 | basque | eu-en | | | 2299 |
| 25 | persian (farsi) | fa-en | | | 3907 |
| 26 | finnish | fi-en | | | 2544 |
| 27 | french | fr-en | | | 2197 |
| 28 | western frisian | fy-en | ✗ | | 1305 |
| 29 | irish | ga-en | | | 2360 |
| 30 | scots | gd-en | | | 1010 |
| 31 | galician | gl-en | | | 1177 |
| 32 | gujarati | gu-en | | | 5588 |
| 33 | hausa | ha-en | ✗ | | 3163 |
| 34 | hebrew | he-en | | | 3109 |
| 35 | hindi | hi-en | | | 4953 |
| 36 | croatian | hr-en | | | 1728 |
| 37 | haitian | ht-en | | ✗ | 675 |
| 38 | hungarian | hu-en | | | 1786 |
| 39 | armenian | hy-en | | | 1733 |
| 40 | indonesian | id-en | | | 1526 |
| 41 | igbo | ig-en | ✗ | ✗ | 1386 |
| 42 | ido | io-en | | ✗ | 3440 |
| 43 | icelandic | is-en | | | 1897 |
| 44 | italian | it-en | | | 2582 |
| 45 | japanese | ja-en | | | 8386 |
| 46 | javanese | jv-en | | | 3351 |
| 47 | georgian | ka-en | | | 4983 |
| 48 | kazakh | kk-en | | | 3213 |
| 49 | central khmer | km-en | ✗ | | 2662 |
| 50 | kannada | kn-en | | | 2238 |
| 51 | korean | ko-en | | | 2691 |
| 52 | kurdish | ku-en | ✗ | | 4001 |
| 53 | kirghiz | ky-en | | | 1052 |
| 54 | latin | la-en | | | 2515 |
| 55 | luxembourgish | lb-en | | ✗ | 1588 |
| 56 | limburgan; limburger; limburgish | li-en | ✗ | ✗ | 1836 |
| 57 | lithuanian | lt-en | | | 2911 |
| 58 | latvian | lv-en | | | 3340 |
| 59 | malagasy | mg-en | | | 3990 |
| 60 | macedonian | mk-en | | | 3607 |
| 61 | malayalam | ml-en | | | 5658 |
| 62 | mongolian | mn-en | ✗ | | 291 |
| 63 | marathi | mr-en | | | 4487 |
| 64 | malay | ms-en | | | 1632 |
| 65 | maltese | mt-en | ✗ | ✗ | 2162 |
| 66 | burmese | my-en | | | 5897 |
| 67 | bokmål, norwegian; norwegian bokmål | nb-en | ✗ | ✗ | 1665 |
| 68 | nepali | ne-en | | | 5587 |
| 69 | dutch | nl-en | | | 1961 |
| 70 | norwegian (nynorsk) | nn-en | | ✗ | 1689 |
| 71 | norwegian (bokmal) | no-en | | | 1345 |
| 72 | occitan | oc-en | | ✗ | 203 |
| 73 | oriya | or-en | ✗ | | 2070 |
| 74 | punjabi | pa-en | | | 3893 |
| 75 | polish | pl-en | | | 2205 |
| 76 | pushto; pashto | ps-en | ✗ | | 2302 |
| 77 | portuguese | pt-en | | | 1987 |
| 78 | romanian | ro-en | | | 1593 |
| 79 | russian | ru-en | | | 3902 |
| 80 | kinyarwanda | rw-en | ✗ | ✗ | 791 |
| 81 | northern sami | se-en | ✗ | ✗ | 336 |
| 82 | sinhala; sinhalese | si-en | ✗ | | 5546 |
| 83 | slovak | sk-en | | | 2322 |
| 84 | slovenian | sl-en | | | 2051 |
| 85 | albanian | sq-en | | | 1881 |
| 86 | serbian | sr-en | | | 1717 |
| 87 | sundanese | su-en | | | 572 |
| 88 | swedish | sv-en | | | 1939 |
| 89 | swahili | sw-en | | | 4234 |
| 90 | tamil | ta-en | | | 5065 |
| 91 | telugu | te-en | | | 5514 |
| 92 | tajik | tg-en | | ✗ | 4434 |
| 93 | thai | th-en | ✗ | | 5853 |
| 94 | turkmen | tk-en | ✗ | ✗ | 576 |
| 95 | tagalog | tl-en | | | 2885 |
| 96 | turkish | tr-en | | | 1945 |
| 97 | tatar | tt-en | | ✗ | 5107 |
| 98 | uighur; uyghur | ug-en | ✗ | | 4529 |
| 99 | ukrainian | uk-en | | | 3544 |
| 100 | urdu | ur-en | | | 4295 |
| 101 | uzbek | uz-en | | | 4675 |
| 102 | vietnamese | vi-en | | | 2260 |
| 103 | walloon | wa-en | ✗ | ✗ | 1181 |
| 104 | xhosa | xh-en | ✗ | | 2597 |
| 105 | yiddish | yi-en | ✗ | | 1856 |
| 106 | yoruba | yo-en | | ✗ | 1082 |
| 107 | chinese (simplified) | zh-en | | | 3302 |
| 108 | zulu | zu-en | ✗ | ✗ | 1801 |
| 109 | serbo-croatian | sh-en | | ✗ | 3423 |
| 110 | asturian | ast-en | | ✗ | 895 |
| 111 | cebuano | ceb-en | | ✗ | 1575 |

Table 9: Parallel language data used in multilingual alignment finetuning.